\documentclass[runningheads]{llncs}

\usepackage[mobile]{eccv}

\usepackage{eccvabbrv}
\usepackage{graphicx}
\usepackage{booktabs}
\usepackage[font=small,labelfont=bf, textfont=it]{caption}
\usepackage{subcaption}
\usepackage{balance}
\usepackage{makecell}
\usepackage{ragged2e}
\usepackage{tabularx}
\usepackage{amsmath}
\usepackage[vlined,ruled]{algorithm2e}
\usepackage[accsupp]{axessibility}
\usepackage[pagebackref,breaklinks,colorlinks,citecolor=eccvblue]{hyperref}

\usepackage{orcidlink}

\begin{document}

\title{MPVO: Motion-Prior based Visual Odometry for PointGoal Navigation}

\author{Sayan Paul \orcidlink{0000-0001-9885-233X} \and
Ruddra dev Roychoudhury \orcidlink{0009-0005-7736-9425} \and
Brojeshwar Bhowmick}

\authorrunning{S. Paul et al.}
\institute{Visual Computing and Embodied AI Lab, TCS Research, India \\
\email{\{p.sayan, ruddra.roychoudhury, b.bhowmick\}@tcs.com}}

\maketitle

\begin{abstract}

Visual odometry (VO) is essential for enabling accurate point-goal navigation of embodied agents in indoor environments where GPS and compass sensors are unreliable and inaccurate. However, traditional VO methods face challenges in wide-baseline scenarios, where fast robot motions and low frames per second (FPS) during inference hinder their performance, leading to drift and catastrophic failures in point-goal navigation. Recent deep-learned VO methods show robust performance but suffer from sample inefficiency during training; hence, they require huge datasets and compute resources. So, we propose a robust and sample-efficient VO pipeline based on motion priors available while an agent is navigating an environment. It consists of a training-free action-prior based geometric VO module that estimates a coarse relative pose which is further consumed as a motion prior by a deep-learned VO model, which finally produces a fine relative pose to be used by the navigation policy. This strategy helps our pipeline achieve up to 2x sample efficiency during training and demonstrates superior accuracy and robustness in point-goal navigation tasks compared to state-of-the-art VO method(s). Realistic indoor environments of the Gibson dataset is used in the AI-Habitat simulator to evaluate the proposed approach using navigation metrics (like success/SPL) and pose metrics (like RPE/ATE). We hope this method further opens a direction of work where motion priors from various sources can be utilized to improve VO estimates and achieve better results in embodied navigation tasks.

\end{abstract}
\section{Introduction}

\begin{figure}[t!]
    \centering
    \includegraphics[width=\linewidth]{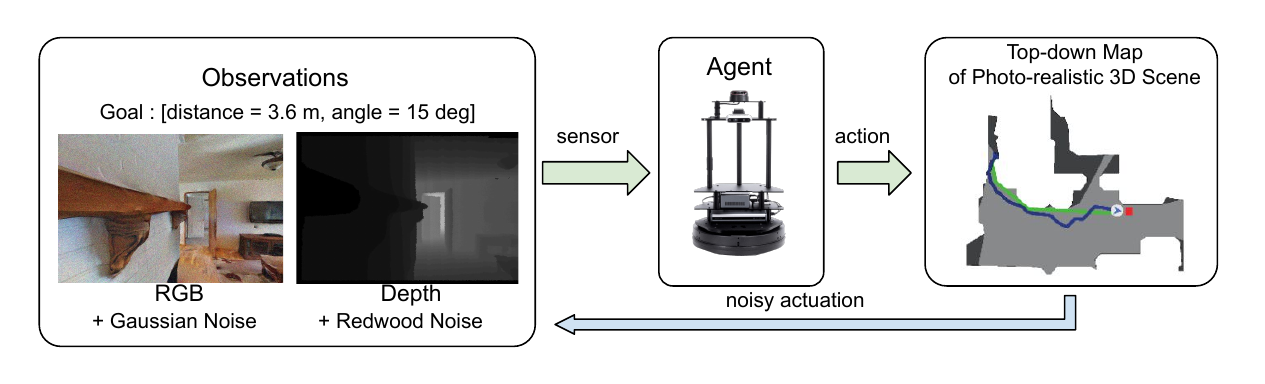}
    \caption{\textbf{Point-Nav Task:} The agent must navigate from its initial location (blue square) to a goal location (red square) specified as goal coordinates w.r.t. its initial location, using only its noisy RGB-D observations and noisy actuation. The agent's and the oracle's path is shown as blue and green lines respectively on the top-down map.}
    \label{fig:pointnav-task}

\end{figure}

Autonomous visual navigation in novel indoor environments is a fundamental skill for robots to  perform further intelligent downstream tasks like finding and retrieving an object, rearranging various stuff, etc. 
This has been the focus of computer vision and robotics researchers for a long time. To bring together the community's efforts and standardize the evaluation framework and metrics, Anderson \textit{et al.} \cite{anderson2018evaluation} proposed the task of PointGoal navigation. In PointNav, (illustrated in \cref{fig:pointnav-task}), the agent or robot is initialized in a previously unseen environment and tasked to reach a goal location specified with respect to its initial location, i.e. go to ($\triangle x$, $\triangle y$). The action space of the agent is discrete and mainly consists of 4 types of action : \texttt{move\_forward}, \texttt{turn\_right}, \texttt{turn\_left} and \texttt{stop} (to end the episode). A pointnav episode is considered successful if the agent stops within a pre-determined distance of the goal location (say for e.g. 0.36 metres) and within the maximum number of time-steps allowed in an episode (say for e.g. 500 time-steps). Apart from success, the agent is evaluated via navigation metrics like SPL \cite{anderson2018evaluation} and SoftSPL \cite{datta2020integrating}.

This point-goal navigation task can either be approached via map-based methods (where the agent simultaneously maps an unexplored area, localizes within it and then plans a path towards the goal) \cite{Chaplot2020LearningTE,karkus2021differentiable} or via recent map-less end-to-end reinforcement learning based methods \cite{datta2020integrating,ZhaoICCV2021,Partsey_2022_CVPR}.
Under the assumption of an ideal scenario, i.e. perfect localization using noiseless GPS+Compass, noiseless egocentric RGB-D sensors and absence of actuation noise; this PointNav task (v1) is fully solved by both map-based and map-less approaches. But the real-world is not ideal, so PointNav v2 came into existence where the agent needs to localize itself (absence of GPS+Compass sensors), has noisy RGB-D observations and noisy actuations. Under this noisy setting, both map-based and map-less approaches need to focus first on accurate localization of the agent, before navigating. Recent map-less  learning based approaches try to solve this by breaking down the task into two parts - learning visual odometry (VO) (for localization) and learning navigation policy (for actions) separately. During inference, this VO model can be used as a drop-in replacement for the GPS+Compass sensor with navigation policies trained using ground-truth pose in simulation.

Now, VO has been studied in Computer/Robot Vision literature for a long-time and many matured solutions exist but this PointNav task definition makes this problem harder to solve. Due to the agent's discrete action space and large motion per action (default: 0.25m forward and 30 deg turns), VO needs to be estimated in a wide camera baseline setting, i.e. the two views are wide-apart and the overlapping region is less. Most of the VO methods, both traditional and learned variants assume availability of frame-pairs with a large overlap or in other words, a narrow baseline setting.
But for practical robot navigation, wide-baseline VO is necessary because situations might arise where the robot motion is fast and/or observation processing or transmission FPS is low. 
Recent learned VO methods used with RL-based navigation policies in map-less approaches \cite{Partsey_2022_CVPR,ZhaoICCV2021} have tried to solve this problem and has achieved robust performance but suffer from sample-inefficiency, embodiment specificity and dataset specificity. It requires huge datasets and compute resources to train such a model and it can't be zero-shot transferred to any other embodiment or dataset. 

So, we propose a robust and sample-efficient novel VO pipeline based on motion priors available while an agent is navigating an environment. We found out from our experiments that simple motion priors like action prior from agent's controller or planner, or coarse pose prior from another geometric pose estimator, etc can help the VO model learn faster using fewer samples than that required by state-of-the-art (SoA) learned methods. This finding not only reduces compute resources needed to train one model but also helpful in scenarios when the target domain data is scarce and costly to collect and curate. Instead of solving for model generalization, we tweaked these models to be more sample-efficient.

We also propose a training-free action-prior based geometric pose estimator or VO module for embodied agents which shows superior performance compared to frame-to-frame VO baselines created using state-of-the-art geometric modules, in standalone evaluation. With this novel geometric VO module, we show that motion priors can be effectively utilized to improve relative pose estimates from sparse feature matching methods. We use this module to estimate a coarse pose which serves as a better motion prior than action prior to train our neural VO model. 

\section{Related Work}

Autonomous visual navigation for indoor robots has been studied for many years \cite{Moravec-1984-15617,durrant-whyte1996localization,gupta2019cognitive}. Recently, due to the advances in deep learning and computer vision, there has been a renewed interest in the use of learning to design navigation policies for a variety of downstream tasks like PointNav \cite{anderson2018evaluation}, ObjectNav \cite{Batra2020ObjectNavRO}, AreaNav \cite{Banerjee2024areanav}, Rearrangement \cite{batra2020rearrangement}, Vision-and-Language based tasks \cite{9846930}, etc.

\subsubsection{Visual Navigation: }
Traditional approaches decompose this task into several sub-tasks such as localization and mapping \cite{ORBSLAM3_TRO,hornung13auro}, followed by planning and control \cite{Hart1968,Rsmann2012TrajectoryMC}. Though these methods can work well when their hyperparameters are hand-tuned properly, errors in one sub-module can propagate to other downstream sub-modules and affect navigation performance adversely. Recent end-to-end learned navigation policies \cite{wijmans2019dd,Partsey_2022_CVPR} alleviate some of these issues and can even outperform traditional methods with sufficient data and training. But they require huge compute resources and sometimes pose problems in generalization.
To circumvent these problems, some modular learned approaches \cite{gupta2019cognitive,Chaplot2020LearningTE,karkus2021differentiable} combine the best of both worlds to retain the benefits of learning the sub-tasks and also the traditional decomposition of the pipeline.

\subsubsection{Visual Odometry: }

VO has been solved using both traditional and learned approaches in the last decade. Most VO methods can be categorized as Sparse or Dense depending on whether the method uses sparse features like keypoints, lines, etc for feature matching or it uses the whole image to determine the optical flow, photo-metric error, etc. Now, sparse methods are generally better suited for wide-baseline settings due to their ability to handle large viewpoint variations, which is our topic of interest. Traditional sparse approaches typically use handcrafted keypoint feature descriptors like SIFT\cite{lowe1999object}, ORB \cite{ORBSLAM3_TRO}, etc followed by correspondence matching and pose estimation using geometric methods. 
Recent learned sparse methods have introduced learned feature descriptors like SuperPoint \cite{detone2018superpoint}, DISK\cite{tyszkiewicz2020disk}, etc and learned correspondence matchers like LightGlue \cite{lindenberger2023lightglue}, SuperGlue\cite{sarlin2020superglue}, etc which perform better than their traditional counterparts. VO methods based on these learned submodules perform well on aggregated relative pose estimation metrics but suffer in trajectory performance due to the agent encountering difficult frame pairs containing featureless plain walls, high depth noise, repetitive patterns, etc. Our proposed geometric VO module leverages motion-priors to constrain and improve its relative pose estimates in such difficult scenarios and perform better than SoA baselines as observed in our experiments.

On the other hand, there has been a surge of dense end-to-end learned VO methods \cite{Wang2017DeepVOTE,zhou2017unsupervised,wang2020tartanvo} in the recent years which directly input image-pairs into CNN based models to regress the relative poses or CNN-RNN models to regress the absolute poses of the camera trajectory. Most of these methods either require narrow camera baseline and/or trained on outdoor autonomous driving datasets which consists less challenging camera trajectories compared to indoor mobile robot scenarios.

In the embodied-AI research community, renewed interest in learning better navigation policies have led to progress in VO models \cite{ZhaoICCV2021,Partsey_2022_CVPR} which take wide-baseline noisy RGB-D image pairs as input and regress the relative pose as output. They exhibit robust performance when used to estimate robot trajectories for pointgoal and other downstream navigation tasks. But they suffer from sample-inefficiency requiring huge compute resources and data. Compared to these methods, our proposed pipeline learns to regress relative pose from frame pairs in a sample-efficient way using the best of the both paradigms - sparse geometric methods and dense learned methods.

\section{Method}

\begin{figure}[t!]
    \centering
    \includegraphics[width=\linewidth]{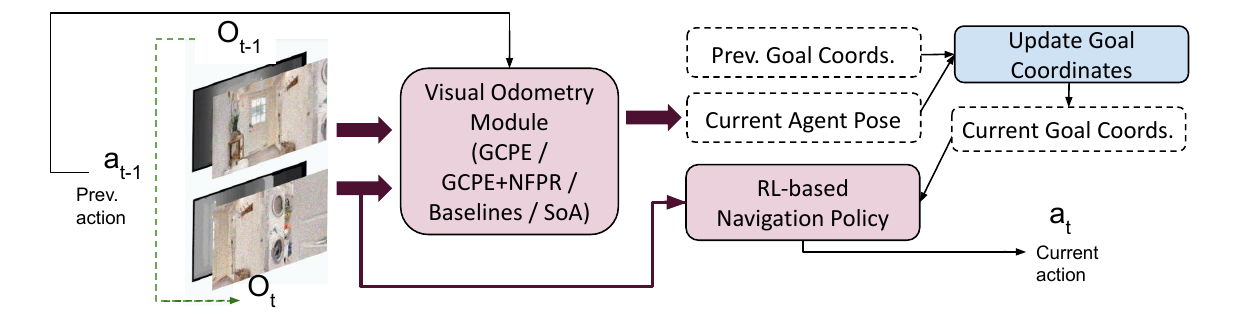}
    \caption{\textbf{Overall Pipeline of the PointNav Agent:} The agent observes $O_t$ upon executing action $a_{t-1}$. The current and previous timestep's observations ($O_t$,$O_{t-1}$) and $a_{t-1}$ are fed into the VO method which outputs the current agent pose after integrating the relative pose estimates till time t. The current observation $O_t$ and the updated goal location w.r.t the current agent pose, is provided to the nav-policy which determines the next action $a_t$.}
    \label{fig:vo_navpolicy_pipeline}

\end{figure}

Our overall pipeline [\cref{fig:vo_navpolicy_pipeline}] consists of two blocks: a visual odometry method (VO) and a navigation policy. This is structured similar to the common practice \cite{Partsey_2022_CVPR,ZhaoICCV2021} of designing pointgoal navigation agents in a map-less setting. The VO method estimates the relative pose between the previous and the current observations, then updates the current goal coordinates using that. The navigation policy determines at each time-step which action to take to reach the goal using this updated goal coordinates and the current observation.

\subsection{Visual Odometry}

Our VO method consists of two key building blocks: one Geometric Coarse Pose Estimator module (GCPE) [\cref{fig:gcpe-module}] and another Neural Fine Pose Regression model (NFPR) [\cref{fig:nfpr-module}]. The pipeline takes as input - two noisy RGB-D frames ($O_{t-1}$ or $O_a$ from previous timestep and $O_t$ or $O_b$ from current timestep) as observed by the agent and an action prior $T_{ap}$ (for e.g. 0.2 m forward, 30 deg right, etc) as provided by the agent's planner or policy. Then, it outputs the relative SE(2) pose $T_{pred}$ ($\triangle x$,$\triangle y$,$\triangle \theta$) between the two RGB-D frames. 

The GCPE module attempts to estimate a coarse relative pose $T_{cp}$ between the two RGB-D frames by performing sparse visual feature point correspondence matching, then sampling candidate relative poses in the SE(2) locality of the given action prior $T_{ap}$ followed by selecting the best scoring candidate pose using a heuristic cost function that re-weighs the 3D-3D point correspondences. It then iteratively refines this pose estimate following the same procedure using the last iteration's pose estimate as it's new prior in the current iteration. Due to noisy depth and lack of good point correspondences in feature-less walls and repetitive textures, this refined pose estimate is still coarse.

So, a neural network model (NFPR) is used to regress the fine pose $T_{pred}$ from the same RGB-D pair and using the above coarse pose estimate and the initial action prior as it's motion priors. This coarse pose estimated by the GCPE module helps improve the overall accuracy and sample-efficiency of the neural model, as compared to directly regressing the fine pose from only RGB-D pair and the action prior.

\begin{algorithm}[t]
\LinesNumbered
\KwIn{
Set of top-m 3D-3D point correspondences ($C^m_{ab}$) from two views, action-prior $T_{ap}$ or 
($x_{ap},y_{ap},\theta_{ap}$) as the initial mean of pose sampling $T^{ps}_{\mu}$, standard-dev of pose sampling $T^{ps}_{\sigma}$ or ($x_{\sigma},y_{\sigma},\theta_{\sigma}$)
}
\KwOut{Coarse Relative Pose $T_{cp}$ ($x_{cp},y_{cp},\theta_{cp}$)}

\SetKwFunction{mpvoestrelpose}{EstimateRelativePose}
\SetKwFunction{mpce}{MotionPriorCorrespondenceWeighting} 
\SetKwFunction{samplerot}{SampleRot}
\SetKwFunction{sampletrans}{SampleTrans}

\SetKwProg{algfunc}{Function}{}{}
\SetFuncSty{small}
\SetCommentSty{small}

\algfunc{
    \mpvoestrelpose{$C^m_{ab}$, $T^{ps}_{\mu}$, $T^{ps}_{\sigma}$}
    }
{
    $j \xleftarrow{}  0$ \;
    $W^{best}_{corr} \xleftarrow{} \phi$ \tcc*[r]{W: correspondence weights}
    \Repeat{ $\frac{Score^{best}_{T_{sample}}-Score^{last.best}_{T_{sample}}}{Score^{best}_{T_{sample}}} > \epsilon_{score}$ }{
    
        \eIf{$j$ is even}{
            
            $T_{samples} \xleftarrow{}$ \samplerot{$T^{ps}_{\mu}$, $T^{ps}_{\sigma}$} \;
        }{
            
            $T_{samples} \xleftarrow{}$ \sampletrans{$T^{ps}_{\mu}$, $T^{ps}_{\sigma}$} \;
        }

        $W_{corr} \xleftarrow{}$ \mpce{$C^m_{ab}, T_{samples}, W^{best}_{corr}$} \;
        $Scores_{T_{samples}} \xleftarrow{} \sum^{c}_{i=1} W_{corr}$\;
        $Score^{best}_{T_{sample}} \xleftarrow{}  \max(Scores_{T_{samples}})$\;
        $idx^{best}_{T_{sample}} \xleftarrow{} \arg\max(Scores_{T_{samples}})$\;
        $W^{best}_{corr} \xleftarrow{} W_{corr}[idx^{best}_{T_{sample}}]$ \;
        $T^{ps}_{\mu} \xleftarrow{} T_{samples}[idx^{best}_{T_{sample}}]$ \;
        
        $T^{ps}_{\sigma} \xleftarrow{} T^{ps}_{\sigma} / 2$ \;
        
        $j \xleftarrow{} j + 1$ \;
    }
    $T_{cp} \xleftarrow{} T^{ps}_{\mu}$ \;
    \Return $T_{cp}$\;
}

\caption{Estimate Pose using Iterative Sampling}
\label{Algo:mpvo}
\end{algorithm}

\subsubsection{Geometric Coarse Pose Estimator (GCPE) :-}

\begin{figure}[t!]
    \centering
    \includegraphics[width=\linewidth]{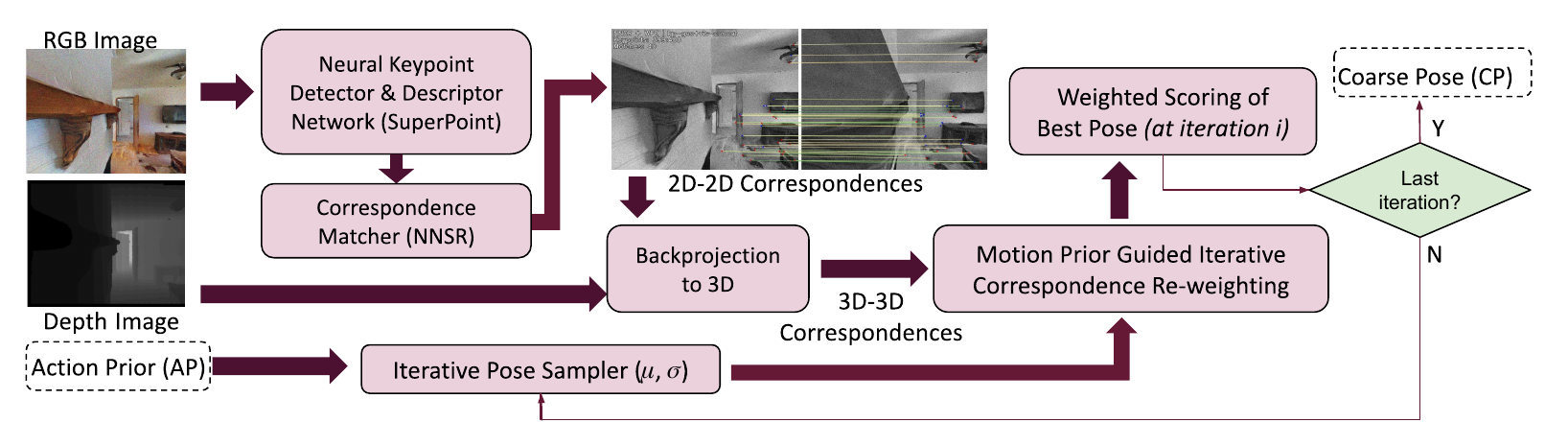}
    \caption{\textbf{Geometric Coarse Pose Estimation (GCPE) module:} this takes an RGB-D pair and action prior pose as input and estimates a coarse pose as output.}
    \label{fig:gcpe-module}

\end{figure}

The GCPE module [\cref{fig:gcpe-module}] first detects and describes sparse visual keypoints in each of the RGB images ($I_a,I_b$) using an off-the-shelf keypoint detector and descriptor (like SuperPoint \cite{detone2018superpoint}, SIFT \cite{lowe1999object}, etc). Then it uses Nearest Neighbour Similarity Ratio (NNSR) (see ratio test in \cite{bian2019unsupervised}) (also known as Lowe's ratio) to match the keypoints $(p_a,p_b)$ and output a set of correspondences $C_{ab}$. Only the top-m correspondences $C^m_{ab}$ according to the scores determined by the ratio test are kept. These 2D point correspondences are back-projected to 3D by using the pixel-aligned depth maps ($D_a,D_b$) and the camera intrinsics $K$.

Then the coarse relative pose between the two views is estimated using \cref{Algo:mpvo}, where a relative pose sampler function generates candidate poses from a normal distribution using the action prior $T_{ap}$ as the mean ($\mu$) and a given large enough standard deviation ($\sigma$) hyper-parameter. The sampler function alternatively samples rotations ($\triangle\theta$) and translations ($\triangle x,\triangle y$) during each iteration. Given the candidate poses $T_{samples}$ and the correspondences $C^m_{ab}$, the Motion-Prior based Correspondence Weighing (MPCW) function weighs each 3D-3D correspondence $C^i_{ab}$ by transforming the points $p^i_a$ from the first view to the second view using the candidate poses, transforming the points $p^i_b$ from the second view to the first view using the inverse of the candidate poses and then calculating the inverse symmetric 3D Endpoint Error (3D-EPE) with their corresponding points $p^i_b$ and $p^i_a$ of the other views respectively. The final weight of each correspondence $W^{j}_{C^i_{ab}}$ is the product of this inverse symmetric 3D-EPE and the best correspondence weights $W^{j-1 (best)}_{C^i_{ab}}$ from the previous iteration. These weights bias the search for the best pose estimate in iteration $j$ around the SE(2) locality of the best pose estimate of iteration $j-1$. For the first iteration, the previous correspondence weights are $\phi$, i.e. not taken into consideration.

\begin{equation}
\small{
W^{j}_{C^i_{ab}} = \frac{W^{j-1 (best)}_{C^i_{ab}}}{(T_{sampled}(p^i_a)-p^i_b)^2 + (T^{-1}_{sampled}(p^i_b)-p^i_a)^2} 
}
\end{equation}

The score of each candidate relative pose is determined by the sum of the weights of all the correspondences. The pose with the highest score is chosen to be the best pose estimate and used as the prior ($\mu$) in next iteration's sampling stage. The standard-deviation ($\sigma$) is halved at the end of each iteration to narrow down the search space. This iterative refining of the coarse pose estimate continues till percentage increase of best pose score with respect to the previous iteration is higher than a given threshold $\epsilon_{score}$. 

\subsubsection{Neural Fine Pose Regression (NFPR) :-}

\begin{figure}[t!]
    \centering
    \includegraphics[width=\linewidth]{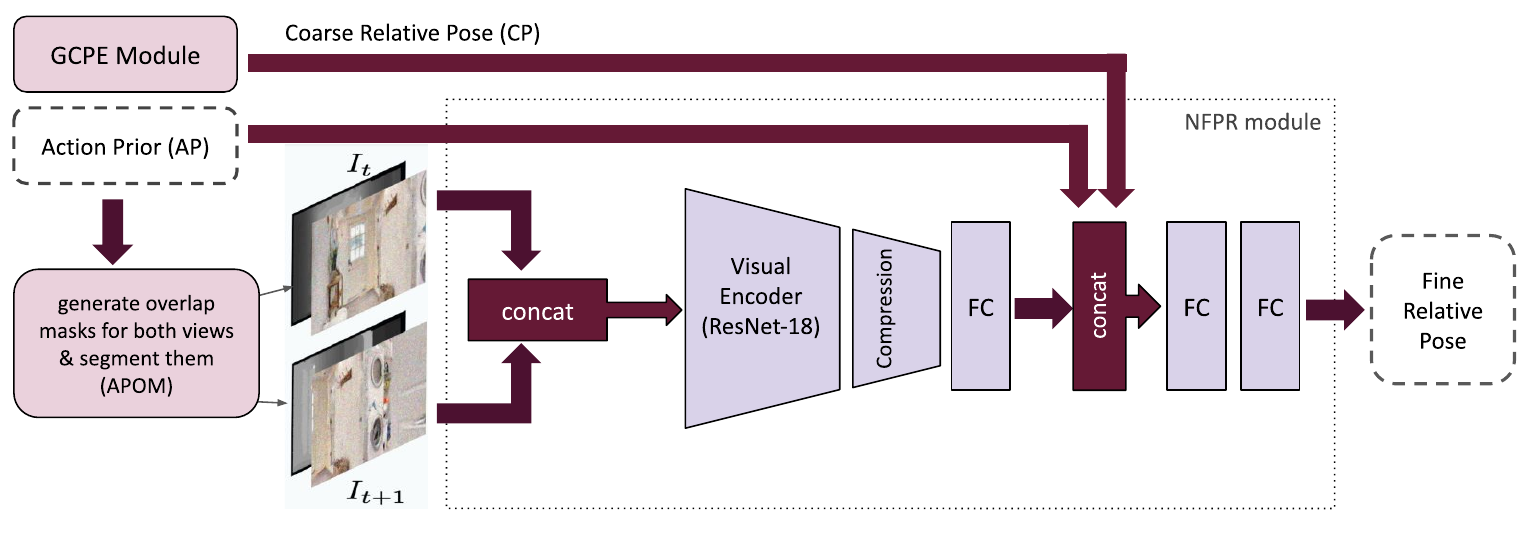}
    \caption{\textbf{Neural Fine Pose Regression (NFPR) module:} this takes an RGB-D pair, action prior pose, coarse prior pose from GCPE module as input and regresses the fine relative pose as output.}
    \label{fig:nfpr-module}

\end{figure}

The NFPR module [\cref{fig:nfpr-module}] is a deep neural network model which takes the RGB-D frame pair, action prior $T_{ap}$ and coarse pose estimate from GCPE module $T_{cp}$ as input and regresses the fine relative pose $T_{pred}$ as output. It consists of a ResNet-18 based visual feature encoder followed by a compression block and a 3-layer MLP pose decoder. Our model architecture is based on \cite{Partsey_2022_CVPR} and we introduce simple modifications which enhance the accuracy and sample-efficiency of the model.

Using the action prior $T_{ap}$ and depth maps of the two views, we generate coarse masks for the overlapping region between the two views. We warp the RGB ($I_a$) of the first view using the depth map ($D_a$) onto the second view using the action prior $T_{ap}$ and warp the RGB ($I_b$) of the second view using the depth map ($D_b$) onto the first view using the inverse of the action prior $T^{-1}_{ap}$ to generate the overlap masks for the first and second views respectively. Then the RGB and Depth images of both views are masked using their corresponding overlap masks, and stacked channel-wise to be fed into the visual encoder of the model. These masked visual observations help focus the model learn dense feature matching in the overlapping regions. We condition the pose decoder MLP by concatenating the raw pose values of the action prior ($\triangle x_{ap}$,$\triangle y_{ap}$,$\triangle \theta_{ap}$) and coarse pose prior ($\triangle x_{cp}$,$\triangle y_{cp}$,$\triangle \theta_{cp}$) to the output of the first FC layer. Note, we do not use the action embedding as mentioned in \cite{Partsey_2022_CVPR}.
The regressed fine pose of the NFPR module is considered as the output of the VO method $\mathbf{T_{pred}}_{t-1}^t$.

\subsection{Navigation Policy}
The relative poses from $\mathbf{T_{pred}}_{0}^1$ to $\mathbf{T_{pred}}_{t-1}^t$ predicted by the VO method till timestep $t$ is integrated to estimate the current agent pose which is further used to update the goal coordinates at the previous timestep $g_{t-1}$ to the current time-step $g_t$. This along-with the current observations $O_t$ is fed to a Reinforcement Learning (RL) based navigation policy to determine the next primitive action (\textit{forward, left, right, stop}) for the agent to take towards the goal location. We adopt the same navigation policy as used by \cite{Partsey_2022_CVPR,wijmans2019dd}, consisting of a half-width ResNet-50 visual encoder and a 2-layer LSTM. Only depth images are given as input, discarding RGB (according to common practise). We use the pre-trained model weights as provided by \cite{Partsey_2022_CVPR} and apply the same pre-processing to the observations as mentioned in the above work. For details, please refer to \cite{Partsey_2022_CVPR}.

\section{Experiments}

\subsection{Dataset}
We use the Habitat Simulator \cite{habitat19iccv}, Gibson Scene dataset \cite{xia2018gibson} and the PointNav v2 task dataset from Habitat-Lab \footnote{https://github.com/facebookresearch/habitat-lab/} for training and evaluation of the overall pointgoal navigation task. For training the Visual Odometry model (NFPR), we generate a static dataset $D = {(O_{t-1}, O_t, T_{ap}, T_{cp}, T_{gt})}$ using an oracle agent (has access to ground-truth map) and shortest path follower policy to unroll the trajectories from which the RGB-D pairs $(O_{t-1}, O_t)$, the action prior $T_{ap}$ and the ground-truth relative pose $T_{gt}$ are uniformly sampled and stored. During this dataset generation phase, the GCPE module is used to compute the coarse motion prior $T_{cp}$ and cache it as part of the dataset sample tuple to be used directly during the training phase. This reduces training time by not having to compute the motion prior for the same observations repeatedly in each epoch. The training dataset (50k to 400k) has been collected by uniformly sampling $20\%$ of the observations from the oracle trajectories of the PointNav v2 task episodes in the 72 scenes of the Gibson 4+ training split. The validation dataset (10k) has been collected by uniformly sampling $75\%$ of the observations from the oracle trajectories of the PointNav v2 task episodes in the 14 scenes of the Gibson 4+ validation split. 

\subsection{Training Details}
The Neural Fine Pose Regression (NFPR) model is trained on the 400k dataset for 50 epochs with batch size of 128, using Adam optimizer with a learning rate of $10^{-4}$ and mean squared error (MSE) loss for both rotation and translation. We save model checkpoints periodically (every epoch) and keep the checkpoint with the lowest validation loss. 

\subsection{Evaluation Details}
To evaluate the performance of the VO module (GCPE and NFPR) in the context of pointgoal navigation, we use the pipeline as shown in \cref{fig:vo_navpolicy_pipeline} and test it against the validation split of the Gibson PointNav v2 task which consists of 994 episodes. For the RL based navigation policy, we use the pretrained model checkpoints as provided by \cite{Partsey_2022_CVPR}. \\ 
The agent is evaluated primarily using 4 different navigation metrics :-
\begin{enumerate}
    \item Success $(S_i)$ : A binary metric which considers an episode to be successful ($S_i = 1$) if the agent stops within 0.36m (2 x agent-radius) of the point-goal, otherwise not ($S_i = 0$).
    \item Success weighted by Path Length (SPL) \cite{anderson2018evaluation}:- 
        $$ 
            \text{SPL } = \frac{1}{N}\sum_{i=1}^{N}S_i \cdot \frac{l_i}{\max(p_i, l_i)}.
        $$ 
        where, $S_i$ is the binary indicator of success, $p_i$ is the agent's path length, and $l_i$ is the shortest path length (geodesic distance) for each episode $i$. And $N$ is the total number of episodes.
    \item Soft-SPL \cite{datta2020integrating}: This metric replaces binary success with the progress towards goal.
        $$ 
            \text{SoftSPL} = \frac{1}{N}\sum_{i=1}^{N}\left(1 - \frac{{d_{T_i}}}{{d_{{0}_i}}}\right)\left(\frac{l_i}{\max(p_i, l_i)}\right). 
        $$ 
        where, $d_{0_i}$ is the initial distance to goal and $d_{T_i}$ is the distance to goal at the end of the episode (on both successes and failures) for each episode $i$.
    \item Distance To Goal $d_g$ : It is the geodesic distance to the goal when the agent issued the "stop" command. 
\end{enumerate}

To understand specifically the VO model's performance, we have also included 3 pose metrics:-  
\begin{enumerate}
    \item Mean Relative Pose Error (RPE) - Rotation \cite{Sturm2012ABF} : This metric computes the rotation error per frame-pair.
    $$ 
        \text{RPE (Rot.)} = \frac{1}{N}\sum_{i=1}^{N} \frac{1}{M}\sum_{j=1}^{M} cos^{-1}(\frac{Trace((\mathbf{R}_{pred}^j)^{-1} \mathbf{R}_{gt}^j)-1}{2}) 
    $$ 
    where, $R_{pred}^j$ and $R_{gt}^j$ are the rotational components of the final predicted relative pose $T_{pred}^j$ and the ground-truth relative pose $T_{gt}^j$ between the frame-pair at time-step $j$ of episode $i$. And $M$ is the total number of timesteps per episode.

    \item Mean Relative Pose Error (RPE) - Translation \cite{Sturm2012ABF}: This metric computes the translation error per frame-pair.
    $$
        \text{RPE (Translation)} = \frac{1}{N}\sum_{i=1}^{N} \frac{1}{M}\sum_{j=1}^{M} ||\mathbf{t}_{pred}^j - \mathbf{t}_{gt}^j||_2
    $$ 
    where, $t_{pred}^j$ and $t_{gt}^j$ are the translational components of the final predicted relative pose $T_{pred}^j$ and the ground-truth relative pose $T_{gt}^j$ between the frame-pair at time-step $j$ of episode $i$.

    \item Mean Absolute Trajectory Error (ATE) - Translation \cite{Sturm2012ABF}: This metric helps to understand the overall drift of the trajectory due to the accumulation of pose errors at each time-step.
    $$
        \text{ATE (Translation)} = \frac{1}{N}\sum_{i=1}^{N} \frac{1}{M}\sum_{j=1}^{M} ||\mathbf{t_{abs}}_{pred}^j - \mathbf{t_{abs}}_{gt}^j||_2
    $$
    where, $\mathbf{t_{abs}}_{pred}^j$ and $\mathbf{t_{abs}}_{gt}^j$ are the translational components of the absolute pose of the predicted trajectory $\mathbf{T_{abs}}_{pred}^j$ and the absolute pose of the ground-truth trajectory $\mathbf{T_{abs}}_{gt}^j$ at time-step $j$ of episode $i$. These absolute poses are computed by integrating frame-wise relative poses from $0$-th timestep to the $j$-th timestep of the agent's trajectory. $$ \mathbf{T_{abs}}^j = \mathbf{T_{abs}}^{j-1} \cdot \mathbf{T_{rel}}^{j-1}_{j} $$

\end{enumerate}

\subsection{Results}
We evaluate and compare the results of our pipeline (GCPE + NFPR) with the state-of-the-art (SoA) VO model from \cite{Partsey_2022_CVPR}. We also compare GCPE module's performance with different geometric pose estimator baselines created using SoA sub-modules. In all the experiments, we have used the same RL-based navigation policy from \cite{Partsey_2022_CVPR} to study the impact of various visual odometry methods on pointgoal navigation.

\subsubsection{GCPE results and baseline comparisons:-}
The Geometric Coarse Pose Estimator is integrated with the Navigation Policy and evaluated on the validation split of the Gibson PointNav v2 task. The hyperparameters of the GCPE module were determined empirically using data from 1 out of the 14 scenes in the Gibson validation split. The standard deviations $x_{\sigma}$, $y_{\sigma}$ and $\theta_{\sigma}$ used by the pose-sampler function were set to $0.06m$, $0.06m$ and $4.0deg$.

\begin{table*}[!htbp]
\scriptsize
    \centering
    \caption{{\textbf{Sub-Modules of the Baselines created to compare with GCPE} (SuperPoint \cite{detone2018superpoint}, LightGlue \cite{lindenberger2023lightglue}, Teaser++ \cite{Yang20tro-teaser}, for Nearest Neighbour Similarity Ratio (NNSR) - see Ratio Test in \cite{elbanani2021unsupervisedrr} and for Randomized Weighted Procrustes (RWP) - see Randomized Optimization in \cite{elbanani2021unsupervisedrr})  }}
    
    \begin{tabular}{|c|c|c|c|}
    \hline
        \textbf{Baseline} & \textbf{Keypoint} & \textbf{Correspondence} & \textbf{Pose} \\ \rule{0pt}{10pt}
        \textbf{} & \textbf{Extractor} & \textbf{Matcher} & \textbf{Estimator} \\ \hline \rule{0pt}{10pt}
        SP-NNSR-TPP & SuperPoint & NNSR & Teaser++ \\  \hline \rule{0pt}{10pt}
        SP-NNSR-RWP & SuperPoint & NNSR & Randomized Weighted Procrustes \\  \hline \rule{0pt}{10pt}
        SP-LG-TPP & SuperPoint & LightGlue & Teaser++ \\  \hline \rule{0pt}{10pt}
        SP-LG-RWP & SuperPoint & LightGlue & Randomized Weighted Procrustes \\  \hline 
        
    \end{tabular}
    
    \label{table_stage1_baseline_defn}

\end{table*}

We created four different baseline methods (\cref{table_stage1_baseline_defn}) to compare with GCPE. All the baselines consists of 3 sub-modules namely the keypoint feature extractor, the correspondence matcher and the pose estimator. The keypoints are detected, described and matched in 2D RGB views from observations $O_{t-1}$ and $O_{t}$. Then these 2D-2D correspondences are back-projected to 3D using depth maps from the same observations. The relative pose is then estimated from these 3D-3D correspondences using the pose estimator sub-module.

\begin{table*}[htbp]
\scriptsize
    \centering
    \caption{\textbf{GCPE module vs SoA geometric pose estimation baselines.} Refer to table:\ref{table_stage1_baseline_defn} for the baseline descriptions. Evaluation results of the pipeline - GCPE/baseline + NavPolicy against the validation split of the Gibson PointNav v2 task dataset.}
    
    \begin{tabularx}{12.2cm}{|c|c|X|X|X|X|X|X|X|} \hline \rule{0pt}{10pt} 

        \textbf{} & \textbf{} & \multicolumn{4}{c|}{\textbf{Navigation Metrics}} & \multicolumn{3}{c|}{\textbf{Pose Metrics}} \\ \hline \rule{0pt}{10pt} 
        
         \textbf{} & \textbf{Method} & \textbf{Success (\%) $\uparrow$} & \textbf{SPL (\%) $\uparrow$} & \textbf{Soft-SPL (\%) $\uparrow$} & \textbf{$d_g$ (cm) $\downarrow$} & \textbf{RPE Rot. MAE (deg) $\downarrow$} & \textbf{RPE Trans. MAE (cm) $\downarrow$} & \textbf{ATE Trans. MAE (cm) $\downarrow$} \\ \hline \rule{0pt}{10pt} 

        1	& sp-nnsr-tpp	& 13.58 & 9.51 & 34.12 & 286.85  & 5.25 & 12.11 & 193.78 \\ \rule{0pt}{10pt}
        2	& sp-nnsr-rwp	& 14.59 & 10.28 & 31.67 & 310.86  & 5.43 & 12.68 & 205.77 \\ \rule{0pt}{10pt}
        3	& sp-lg-tpp	& 18.61 & 13.68 & 37.72 & 245.07 & 2.56 & 9.56 & 182.06 \\ \rule{0pt}{10pt}
        4	& sp-lg-rwp  & 28.77 & 20.88 & 46.20 & 188.18 & 1.71 & 6.28 & 107.29 \\ \hline \rule{0pt}{10pt}
        5	& mpvo-gcpe (ours) & 35.31 & 26.19 & 57.03 & 120.56 & 1.42 & 4.54 & 58.80 \\ \hline

    \end{tabularx}
    
    \label{table_stage1}

\end{table*}

We selected Superpoint \cite{detone2018superpoint} as the keypoint extractor for all the baselines and our GCPE module as it is one of the highly accurate learned keypoint feature descriptors with open-source pre-trained model weights available. Superpoint performs better than traditional hand-crafted keypoint feature descriptors like SIFT, ORB, etc as shown in their original work. For the feature matcher, we selected one classical method - Nearest Neighbour Similarity Ratio (NNSR) (refer to Ratio test in \cite{elbanani2021unsupervisedrr}) and another learned SoA method - LightGlue \cite{lindenberger2023lightglue}. Both the methods assign weights to the correspondences between 0 and 1 where higher the weight, more the probability of the correspondence being an inlier. As we can observe from \cref{table_stage1}, LightGlue performs better than NNSR for the same keypoint extractor and pose estimator combination. This is due to its superior outlier correspondence filtering. For the pose estimator, we selected a modified variant of the classical Weighted Procrustes method - Randomized Weighted Procrustes (RWP) (refer to randomized optimization in \cite{elbanani2021unsupervisedrr}) and another SoA robust pose estimation method - Teaser++ \cite{Yang20tro-teaser}. We observe that RWP performs better than Teaser++ for the same keypoint extractor and matcher combination. We hypothesize that this might be due to the fact that Teaser++'s algorithm doesn't use the correspondence weights and prunes the outliers based on hard thresholds. Whereas RWP effectively uses the weights from either NNSR or LightGlue for inlier/outlier soft-assignment and gives an overall better pose estimate.

As we can observe from \cref{table_stage1}, GCPE succeeds in around 35\% episodes, performs better than the baselines created using off-the-shelf SoA submodules and shows a drastic improvement in ATE. This can be attributed to the fact that none of the other baselines bias their pose estimation using the action prior of the agent. So, in case of the baselines, this unbiased pose estimation results in large pose errors when the agent encounters difficult frame-pairs containing featureless walls, high depth noise, etc which affects the correspondence matching step thereby increasing the number of outliers. These errors in some of the frame-pairs accumulate over time and causes huge trajectory drift as evidenced by ATE. On the other hand, GCPE safeguards the agent's pose estimates by performing iterative re-weighting of correspondences (or outlier correspondences filtering, in other words) using the action prior as an initial estimate and converging towards the coarse pose estimate.

\subsubsection{Our method (GCPE+NFPR) results and comparisons:-} 
As evident from \cref{table_dataset_ablations}, our method comprising of the GCPE module and our best performing NFPR model (APOM-APCP) consistently performs better than SoA method \cite{Partsey_2022_CVPR} for the same dataset size and also achieves nearly same navigation success using approximately half the amount of data, i.e. upto 2x more sample efficient than SoA.
Our best performing model (APOM-APCP) trained on 400k reports much lower ATE than that of SoA \cite{Partsey_2022_CVPR} which indicates lesser outlier pose estimates (frame-pairs having large pose errors compared to that of rest of the frame-pairs) along the trajectory. This superiority can be attributed to the various simple yet effective modifications applied to the base-model, especially the inclusion of motion priors from the GCPE module and the action-prior based overlap masks for the RGBD observations.

\begin{table}[!ht]
\scriptsize
    \centering
    \caption{\textbf{Our Method vs SoA Method} - on increasing the dataset size. Success(\%) has been rounded off to the nearest integer. Notice that our method is more sample-efficient (upto 2x) than the SoA method.}
    
    \begin{tabular}{|c|c|c|}
    \hline
        \textbf{} & \multicolumn{2}{c|}{\textbf{Navigation Success (\%)}} \\ \hline \rule{0pt}{10pt}
        \textbf{Dataset Size} & \textbf{Ours (APOM-APCP)} & \textbf{Partsey etal. (SoA)} \\ \hline \rule{0pt}{10pt}
        50k & 63.0 & 35.0 \\  \hline \rule{0pt}{10pt}
        100k & 66.0 & 60.0 \\ \hline \rule{0pt}{10pt}
        200k & 72.0 & 65.0 \\ \hline \rule{0pt}{10pt}
        400k & 78.0 & 72.0 \\ \hline
    \end{tabular}
    
    \label{table_dataset_ablations}

\end{table}

\subsubsection{NFPR Ablations [\cref{table_main}] :-}
To gain insights about which modifications contribute to our model's superiority, we ablate the three major modifications in various combinations and evaluate agent's performance against the Gibson PointNav v2 validation split. We also perform the ablations at varying dataset sizes (50k, 200k and 400k) to confirm whether the benefits remain consistent or not.

\begin{table*}[htbp]
\scriptsize
    \centering
    \caption{\textbf{NFPR Model Ablations and Comparison with SoA} \cite{Partsey_2022_CVPR}. Evaluation results of the overall pipeline - GCPE + NFPR + NavPolicy against the validation split of the Gibson PointNav v2 task dataset. All the ablations are run 4 times each using a different seed and then averaged. }
    
    \begin{tabularx}{12.2cm}{|c|X|c|c|c|X|X|X|X|X|X|X|} \hline \rule{0pt}{10pt} 

        \textbf{} & \textbf{} & \multicolumn{3}{c|}{\textbf{Model}} & \multicolumn{4}{c|}{\textbf{Navigation Metrics}} & \multicolumn{3}{c|}{\textbf{Pose Metrics}} \\ \hline \rule{0pt}{10pt} 
        
         \textbf{} & \textbf{Dataset} \newline \textbf{Samples (K)} & \textbf{AP} & \textbf{CP} & \textbf{APOM} & \textbf{Success (\%) $\uparrow$} & \textbf{SPL (\%) $\uparrow$} & \textbf{Soft-SPL (\%) $\uparrow$} & \textbf{$d_g$ (cm) $\downarrow$} & \textbf{RPE Rot. MAE (deg) $\downarrow$} & \textbf{RPE Trans. MAE (cm) $\downarrow$} & \textbf{ATE Trans. MAE (cm) $\downarrow$} \\ \hline \rule{0pt}{10pt} 
        1 & 50 & \textbf{*} & ~ & ~ & 37.27 & 28.90 & 63.20 & 119.71 & 1.45 & 3.86 & 46.37 \\ \rule{0pt}{10pt}
        2 & 50 & ~ & \textbf{*} & ~ & 57.79 & 45.11 & 67.87 & 76.01 & 0.89 & 2.51 & 28.57 \\ \rule{0pt}{10pt}
        3 & 50 & \textbf{*} & \textbf{*} & ~ & 54.70 & 42.36 & 66.32 & 95.78 & 0.94 & 2.55 & 31.75 \\ \rule{0pt}{10pt}
        4 & 50 & \textbf{*} & ~ & \textbf{*} & 45.85 & 35.94 & 66.29 & 101.8 & 0.99 & 2.93 & 37.86 \\ \rule{0pt}{10pt}
        5 & 50 & \textbf{*} & \textbf{*} & \textbf{*} & 63.40 & 52.44 & 69.74 & 71.15 & 0.75 & 2.21 & 22.54 \\ \hline \rule{0pt}{10pt}
        8 & 200 & \textbf{*} & \textbf{*} & ~ & 69.11 & 53.95 & 70.01 & 55.12 & 0.64 & 2.07 & 20.18 \\ \rule{0pt}{10pt}
        9 & 200 & \textbf{*} & ~ & \textbf{*} & 67.51 & 52.96 & 69.81 & 74.31 & 0.65 & 2.06 & 23.21 \\ \rule{0pt}{10pt}
        10 & 200 & \textbf{*} & \textbf{*} & \textbf{*} & 72.18 & 55.04 & 69.68 & 56.16 & 0.57 & 1.69 & 20.62 \\ \hline \rule{0pt}{10pt}
        11 & 400 & \textbf{*} & \textbf{*} & \textbf{*} & 78.21 & 61.01 & 71.87 & 45.06 & 0.51 & 1.52 & 15.51 \\
        \hline \rule{0pt}{10pt}
        12 & 400 & \multicolumn{3}{c|}{Partsey.etal(SoA)} & 72.23 & 55.99 & 70.34 & 59.12 & 0.56 & 1.80 & 19.09 \\ \hline
    
    \end{tabularx}
    
    \label{table_main}

\end{table*}

\begin{itemize}
    \item \textbf{Effect of Coarse Pose Prior (CP):} As evident from \cref{table_main}, the inclusion of CP benefits the model a lot as compared to only action-prior (AP) and also with overlap masks (APOM), hence the need for GCPE or any other coarse pose estimator is established. This is due to the fact that CP is much more accurate than AP, so it helps the model learns the residual $\triangle pose$ needed to make it close enough to the ground-truth (GT) pose, thereby converging in the right direction and decreasing the training loss. 

    \item \textbf{Effect of Action Prior (AP):} Using only coarse motion-priors (CP) from a visual pose estimation module (like GCPE) to train the NFPR model may lead to loss plateauing upon prolonged training. This can be attributed to the fact that CP being a stochastic motion prior can hinder the training loss converge to the global minimum. We found out that simply feeding the action prior along-with the coarse motion prior to the pose decoder block provides a deterministic biasing effect to the model optimization. This helps the model learn better for a longer time on larger datasets. It can be noticed that CP alone performs better than AP+CP on 50k dataset but this gains diminish and the above problems arise when increasing dataset size. Note, we don't use action embedding as in \cite{Partsey_2022_CVPR}, instead prefer to use action priors (AP) directly as we found the later to be more effective in most cases (check row-1 of \cref{table_main} vs row-1 (Partsey. etal which uses action embedding) of \cref{table_dataset_ablations}).

    \item \textbf{Effect of Action Prior based Overlap Masks (APOM):} In case of the usual unmasked RGB-D inputs, the model needs to learn and figure out the overlapping region between the two views in order to predict an accurate relative pose. This is a hard problem for simple CNNs and is usually approached by using attention mechanism of transformers. In order to keep a lightweight CNN-based model and yet help the model focus or attend to the overlapping region of the two views, we explicitly mask the RGB-D inputs by re-projecting one view to another and vice-versa using the action prior pose. This hard-attention helps the model perform better as evident from \cref{table_main}. Note, we could have also used coarse motion prior (CP) to generate the overlap masks but we observed from our experiments that APOM performs better than CPOM due to the highly stochastic nature of CP which hinders the model from learning consistent patterns.  
    
\end{itemize}

\section{Conclusion}

In this work, we propose MPVO, a robust and sample-efficient VO pipeline for use in pointgoal navigation of embodied agents. It's based on the effective utilization of motion priors available during agent navigation. We have demonstrated it using action priors from the agent planner, but other motion priors like wheel odometry, IMU, etc can also be utilized to improve its efficacy further. We have conducted extensive experiments to show that MPVO performs better and is more sample-efficient (upto 2x) than SoA. Our training-free geometric coarse pose estimator (GCPE) also performs better than SoA baselines in standalone evaluation. Though we have shown our VO pipeline's usage in the context of pointgoal navigation, this can be used in other embodied navigation tasks such as ObjectNav, Rearrangement, etc. We hope that this work motivates further research in utilizing motion priors to improve VO estimates for navigation tasks.

\bibliographystyle{splncs04}

\begin{thebibliography}{10}
\providecommand{\url}[1]{\texttt{#1}}
\providecommand{\urlprefix}{URL }
\providecommand{\doi}[1]{https://doi.org/#1}

\bibitem{anderson2018evaluation}
Anderson, P., Chang, A., Chaplot, D.S., Dosovitskiy, A., Gupta, S., Koltun, V., Kosecka, J., Malik, J., Mottaghi, R., Savva, M., Zamir, A.R.: On evaluation of embodied navigation agents (2018)

\bibitem{Banerjee2024areanav}
Banerjee, S., Paul, S., Roychoudhury, R., Bhattacharya, A., Sarkar, C., Sau, A., Pramanick, P., Bhowmick, B.: Teledrive: An embodied ai based telepresence system. Journal of Intelligent {\&} Robotic Systems  \textbf{110}(3), ~96 (Jul 2024). \doi{10.1007/s10846-024-02124-0}

\bibitem{batra2020rearrangement}
Batra, D., Chang, A.X., Chernova, S., Davison, A.J., Deng, J., Koltun, V., Levine, S., Malik, J., Mordatch, I., Mottaghi, R., Savva, M., Su, H.: Rearrangement: A challenge for embodied ai (2020)

\bibitem{Batra2020ObjectNavRO}
Batra, D., Gokaslan, A., Kembhavi, A., Maksymets, O., Mottaghi, R., Savva, M., Toshev, A., Wijmans, E.: Objectnav revisited: On evaluation of embodied agents navigating to objects. ArXiv  \textbf{abs/2006.13171} (2020)

\bibitem{bian2019unsupervised}
Bian, J., Li, Z., Wang, N., Zhan, H., Shen, C., Cheng, M.M., Reid, I.: Unsupervised scale-consistent depth and ego-motion learning from monocular video. Advances in neural information processing systems  \textbf{32} (2019)

\bibitem{ORBSLAM3_TRO}
Campos, C., Elvira, R., Gomez, J.J., Montiel, J.M.M., Tardos, J.D.: {ORB-SLAM3}: An accurate open-source library for visual, visual-inertial and multi-map {SLAM}. IEEE Transactions on Robotics  \textbf{37}(6),  1874--1890 (2021)

\bibitem{Chaplot2020LearningTE}
Chaplot, D.S., Gandhi, D., Gupta, S., Gupta, A.K., Salakhutdinov, R.: Learning to explore using active neural slam. ArXiv  \textbf{abs/2004.05155} (2020)

\bibitem{datta2020integrating}
Datta, S., Maksymets, O., Hoffman, J., Lee, S., Batra, D., Parikh, D.: Integrating egocentric localization for more realistic point-goal navigation agents (2020)

\bibitem{detone2018superpoint}
DeTone, D., Malisiewicz, T., Rabinovich, A.: Superpoint: Self-supervised interest point detection and description (2018)

\bibitem{durrant-whyte1996localization}
Durrant-Whyte, H., Rye, D., Nebot, E.: Localization of autonomous guided vehicles. In: Robotics Research (1996)

\bibitem{elbanani2021unsupervisedrr}
El~Banani, M., Gao, L., Johnson, J.: {UnsupervisedR\&R: Unsupervised Pointcloud Registration via Differentiable Rendering}. In: CVPR (2021)

\bibitem{gupta2019cognitive}
Gupta, S., Tolani, V., Davidson, J., Levine, S., Sukthankar, R., Malik, J.: Cognitive mapping and planning for visual navigation (2019)

\bibitem{Hart1968}
Hart, P., Nilsson, N., Raphael, B.: A formal basis for the heuristic determination of minimum cost paths. {IEEE} Transactions on Systems Science and Cybernetics  \textbf{4}(2),  100--107 (1968). \doi{10.1109/tssc.1968.300136}

\bibitem{hornung13auro}
Hornung, A., Wurm, K.M., Bennewitz, M., Stachniss, C., Burgard, W.: {OctoMap}: An efficient probabilistic {3D} mapping framework based on octrees. Autonomous Robots  (2013)

\bibitem{karkus2021differentiable}
Karkus, P., Cai, S., Hsu, D.: Differentiable slam-net: Learning particle slam for visual navigation (2021)

\bibitem{lindenberger2023lightglue}
Lindenberger, P., Sarlin, P.E., Pollefeys, M.: Lightglue: Local feature matching at light speed (2023)

\bibitem{lowe1999object}
Lowe, D.G.: Object recognition from local scale-invariant features. In: Proceedings of the seventh IEEE international conference on computer vision. vol.~2, pp. 1150--1157. Ieee (1999)

\bibitem{Moravec-1984-15617}
Moravec, H.: Locomotion, vision and intelligence. In: Brady, M., Paul, R. (eds.) Proceedings of 2nd International Symposium on Robotics Research (ISRR '84). pp. 215 -- 224. MIT Press (August 1984)

\bibitem{Partsey_2022_CVPR}
Partsey, R., Wijmans, E., Yokoyama, N., Dobosevych, O., Batra, D., Maksymets, O.: Is mapping necessary for realistic pointgoal navigation? In: Proceedings of the IEEE/CVF Conference on Computer Vision and Pattern Recognition (CVPR). pp. 17232--17241 (June 2022)

\bibitem{9846930}
Pramanick, P., Sarkar, C., Paul, S., Roychoudhury, R.d., Bhowmick, B.: Doro: Disambiguation of referred object for embodied agents. IEEE Robotics and Automation Letters  \textbf{7}(4),  10826--10833 (2022). \doi{10.1109/LRA.2022.3195198}

\bibitem{Rsmann2012TrajectoryMC}
R{\"o}smann, C., Feiten, W., W{\"o}sch, T., Hoffmann, F., Bertram, T.: Trajectory modification considering dynamic constraints of autonomous robots. In: German Conference on Robotics (2012)

\bibitem{sarlin2020superglue}
Sarlin, P.E., DeTone, D., Malisiewicz, T., Rabinovich, A.: Superglue: Learning feature matching with graph neural networks (2020)

\bibitem{habitat19iccv}
Savva, M., Kadian, A., Maksymets, O., Zhao, Y., Wijmans, E., Jain, B., Straub, J., Liu, J., Koltun, V., Malik, J., Parikh, D., Batra, D.: Habitat: {A} {P}latform for {E}mbodied {AI} {R}esearch. In: Proceedings of the IEEE/CVF International Conference on Computer Vision (ICCV) (2019)

\bibitem{Sturm2012ABF}
Sturm, J., Engelhard, N., Endres, F., Burgard, W., Cremers, D.: A benchmark for the evaluation of rgb-d slam systems. 2012 IEEE/RSJ International Conference on Intelligent Robots and Systems pp. 573--580 (2012)

\bibitem{tyszkiewicz2020disk}
Tyszkiewicz, M.J., Fua, P., Trulls, E.: Disk: Learning local features with policy gradient (2020)

\bibitem{Wang2017DeepVOTE}
Wang, S., Clark, R., Wen, H., Trigoni, A.: Deepvo: Towards end-to-end visual odometry with deep recurrent convolutional neural networks. 2017 IEEE International Conference on Robotics and Automation (ICRA)  (2017)

\bibitem{wang2020tartanvo}
Wang, W., Hu, Y., Scherer, S.: Tartanvo: A generalizable learning-based vo (2020)

\bibitem{wijmans2019dd}
Wijmans, E., Kadian, A., Morcos, A., Lee, S., Essa, I., Parikh, D., Savva, M., Batra, D.: Dd-ppo: Learning near-perfect pointgoal navigators from 2.5 billion frames. arXiv preprint arXiv:1911.00357  (2019)

\bibitem{xia2018gibson}
{Xia}, F., {Zamir}, A.R., {He}, Z., {Sax}, A., {Malik}, J., {Savarese}, S.: Gibson env: Real-world perception for embodied agents. In: 2018 IEEE/CVF Conference on Computer Vision and Pattern Recognition. pp. 9068--9079 (2018)

\bibitem{Yang20tro-teaser}
Yang, H., Shi, J., Carlone, L.: {TEASER: Fast and Certifiable Point Cloud Registration}. {IEEE} Trans. Robotics  (2020)

\bibitem{ZhaoICCV2021}
Zhao, X., Agrawal, H., Batra, D., Schwing, A.: {The Surprising Effectiveness of Visual Odometry Techniques for Embodied PointGoal Navigation}. In: Proc. ICCV (2021)

\bibitem{zhou2017unsupervised}
Zhou, T., Brown, M., Snavely, N., Lowe, D.G.: Unsupervised learning of depth and ego-motion from video (2017)

\end{thebibliography}

\end{document}